# The use of a humanoid robot for older people with dementia in aged care facilities

Dongjun Wu, Lihui Pu, Jun Jo, Rene Hexel, Wendy Moyle

*Abstract*— **This paper presents an interdisciplinary PhD project using a humanoid robot to encourage interactive activities for people with dementia living in two aged care facilities. The aim of the project was to develop software and use technologies to achieve successful robot-led engagement with older people with dementia. This paper outlines the qualitative findings from the project's feasibility stage. The researcher's observations, the participants' attitudes and the feedback from carers are presented and discussed.**

## I. INTRODUCTION

Dementia is a disorder that affects cognitive function and is common in older people [1]. There is no cure for dementia and alongside an aging population, the burden of dementia continues to increase for people with dementia, their carers and the healthcare system [2, 3]. According to a growing body of research, encouraging activity engagement is important for people with dementia to maintain daily functioning and social skills [4]. In addition, participating in activities may help people with dementia achieve enjoyment, satisfaction, and quality of life [5]. This project provides an innovative approach to introduce 'Adam' a humanoid robot and to offer interactive activities in authentic contexts. In addition, the project aims to explore the feasibility of allowing humanoid robots to encourage activity engagement without human facilitation. This paper describes using a humanoid robot in two aged care facilities in Brisbane, Australia.

## II. METHOD

### A. Robot programming

The control logic for a robot called "Adam" was programmed and implemented in a desktop humanoid robot product (Alpha Mini) to provide guidance and encourage participants to engage in a series of activities (Fig.1). These activities are social interaction activities such as verbal communication, following movements (i.e., dancing) and finding items (i.e., following instructions to recognize a toy placed in a basket). An iPad screen was placed beside the robot to display Adam's speech and some hints for participants to work with (Fig.1).

### B. Intervention and Participants

Eleven (10 female) older adults (65 years or older) with mild cognitive impairment participated in the robot activity three times a week for five weeks. The robot session was individual and conducted in a private room (bedroom or activity room). During the 10-15-minute individual interaction session, the researcher moved away from the participant to an area where they could observe and were not in the direct vision of the participant (Fig.1). Participants were left to interact with the robot freely. After the session, the researcher returned and guided the participant to say goodbye to the robot. This study was approved by Griffith University Human Research Ethics Committee (GU Ref No: 2022/392). All participants provided written informed consent before the study.

## III. FINDINGS

Collectively, we observed a high level of activity engagement from participants. The study findings demonstrate the promise of a fully automatic robot driven by different types of technologies in encouraging activity engagement for people with dementia.

### A. Observation of the robot session

Most participants got along well with Adam and could engage in interactive activities independently. Participants 001, 002, 004, 008, 010, and 011 enjoyed the interactive game and expressed happiness with Adam's reaction when they identified different toys. Sometimes Adam provided a wrong answer to their selection (i.e., he recognized the camel as a deer), and participants appeared to enjoy the interaction but not the answer. Over half of the participants 001, 002, 004, 009, 010, and 011 expressed they loved Adam's emotional response (i.e., Adam's eyes) during or after the robot session. Participants 001, 003, 006, 007, 008, 010, and 011, enthusiastically followed Adam's dancing. In addition, Adam's positive words, such as "well done" or "You did very well," successfully maintained the interaction between Adam and the participants.

During the intervention period, we observed several robot failures, such as the program crashing, network delay, a blank screen display, wrong object detection, and the failure of speech detection. Although participants sometimes got confused or anxious about the situation, such technical issues did not cause adverse effects to participants in either robot session. In addition, Adam could handle most situations (i.e., wrong object detection and the failure of speech detection) independently and maintained interaction with participants. Only the program crashing, blank screen display, and network issues required the researcher's facilitation.

### B. Participants' attitude

All participants attended an interview after the last robot session. Overall, participants stated they were satisfied with the robot session regarding robot interaction and activity

Dongjun Wu, Lihui Pu, and Wendy Moyle are with the School of Nursing and Midwifery, Menzies Health Institute Queensland (e-mail: dongjun.wu@griffithuni.edu.au, {l.pu, w.moyle}@griffith.edu.au).
Jun Jo and Rene Hexel are with the School of Information and Communication Technology, (e-mail: {j.jo, r.hexel}@griffith.edu.au)).
All authors are with Griffith University, Brisbane, Australia.

design. Participants 001, 004, 008, and 010 reported feeling happy when interacting with the robot. Although some participants, such as 005, 007, and 011 needed help remembering the details of the interactive activities, all participants found that Adam encouraged them to participate. In addition, most of the participants treated Adam as a child or human and said they would like to introduce Adam to their family.

Moreover, all participants found it helpful to display the robot's speech and hints on the screen. Participant 009 also mentioned that sometimes she was not clear about Adam's speech, but she could see it on the screen. Participants 001, 004, 005, 010, and 011 would like to play more games with Adam in the future. More robot features, such as singing and music, were suggested by three participants, 005, 008 and 009. One participant, 002, asked that Adam change his voice in the future.

*C. Feedback from carers at the aged care facility*

The response from nurses and carers was positive and supportive. Some carers reported that residents who participated in the research felt happy to share their robot session with neighbours and carers. One nurse mentioned that participant 006 enjoyed staying with Adam and expressed her happiness for participating in the research.

## IV. DISCUSSION AND FUTURE WORK

This study explores the feasibility of robots to encourage activity engagement for people with dementia who live in an aged care facility. Our findings demonstrate a feasible and acceptable robot-led intervention without human facilitation and successfully encourage people with dementia to participate in interactive activities.

*A. The promise of interactive activity led by Adam*

Firstly, we observed that robot activities driven by technologies could maintain the interaction between Adam and the participants. Although we observed that participants got confused during the robot session, the encouraging words from Adam and the screen display facilitated participants to work out a way to interact with Adam. Second, the positive attitude from all participants reveals that people with dementia were accepting of Adam, a human-like robot. Moreover, our findings suggested that Adam's human-like personality (i.e., physical embodiment, behaviors, and facial expression) may facilitate the engagement of robot activities such as dancing together and verbal conversation and potentially influence some participants to treat Adam as a human or child. Ultimately, the robot-led intervention design did not cause additional workload to nurses and carers when delivering the robot activity. The supportive response from carers revealed that the robot activity may help them provide daily activity to people with dementia.

*B. Limitations and future work*

Our findings demonstrate the feasibility of robot-led activity for people with dementia in aged care facilities. However, more research with a rigorous study design (i.e., larger sample size and quantitative data) is needed to evaluate the effect of such robot activities. In addition, technical issues such as program crashes, screen display failures and wrong detection confused people with dementia during the robot activity. Moreover, the involvement of researchers in solving technical issues may influence the result of such a robot activity. Therefore, the robot program requires further improvement to reduce technical issues (i.e., program crashing) and to minimize the researcher's influence. Ultimately, ethical issues should be a concern in future research as our result revealed that people with dementia would likely treat a human-like robot as a human [6].

ACKNOWLEDGMENT

We acknowledge the support of the Southern Cross Age Care management team for supporting our research and the people with dementia and their families.

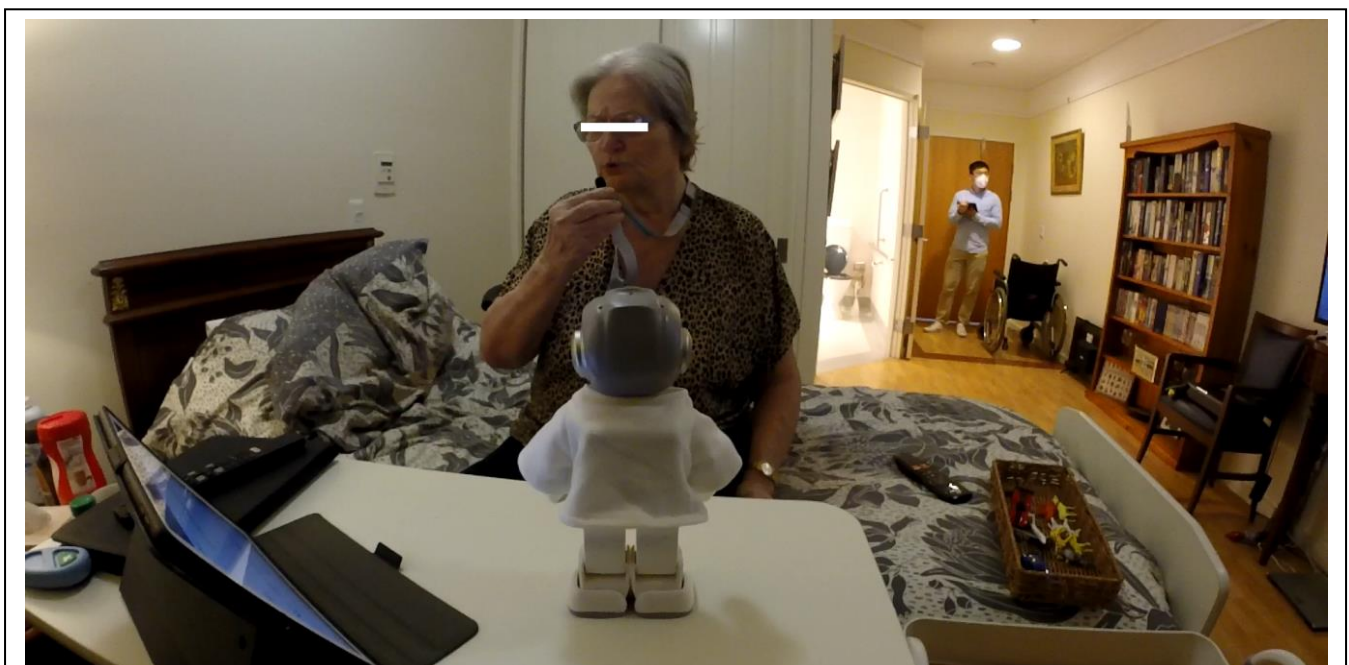

Figure 1. The demonstration of robot activity. *The participant interacts with the robot independently.*

REFERENCES

[1] World Health Organization. "Dementia." https://www.who.int/health-topics/dementia#tab=tab_2 (accessed May 8, 2023.

[2] United Nations, "Department of Economic and Social Affairs, Population Division (2019). World Population," 2019, https://population.un.org/wpp/Publications/Files/WPP2019_Highlights.pdf.

[3] World Health Organization. "Dementia Fact Sheets." https://www.who.int/news-room/fact-sheets/detail/dementia (accessed May 25, 2023).

[4] N. Mavridis, "A review of verbal and non-verbal human–robot interactive communication," *Robot Auton Syst,* vol. 63, pp. 22-35, 2015/01/01/ 2015, https://doi.org/10.1016/j.robot.2014.09.031.

[5] M. Ghafurian, J. Hoey, and K. Dautenhahn, "Social Robots for the Care of Persons with Dementia: A Systematic Review," *J Hum-Robot Interact,* vol. 10, no. 4, p. Article 41, 2021, https://doi.org/10.1145/3469653.

[6] R. Etemad-Sajadi, A. Soussan, and T. Schöpfer, "How Ethical Issues Raised by Human–Robot Interaction can Impact the Intention to use the Robot?," *Int J Soc Robot,* vol. 14, no. 4, pp. 1103-1115, 2022/06/01 2022, https://doi.org/10.1007/s12369-021-00857-8.